# Improving the Scalability of Optimal Bayesian Network Learning with External-Memory Frontier Breadth-First Branch and Bound Search

**Brandon Malone** and **Changhe Yuan** and **Eric A. Hansen** and **Susan Bridges**
Department of Computer Science and Engineering
Mississippi State University
Mississippi State, MS 39762
bm542@msstate.edu, {cyuan, hansen, bridgess}@cse.msstate.edu

## Abstract

Previous work has shown that the problem of learning the optimal structure of a Bayesian network can be formulated as a shortest path finding problem in a graph and solved using A* search. In this paper, we improve the scalability of this approach by developing a memory-efficient heuristic search algorithm for learning the structure of a Bayesian network. Instead of using A*, we propose a frontier breadth-first branch and bound search that leverages the layered structure of the search graph of this problem so that no more than two layers of the graph, plus solution reconstruction information, need to be stored in memory at a time. To further improve scalability, the algorithm stores most of the graph in external memory, such as hard disk, when it does not fit in RAM. Experimental results show that the resulting algorithm solves significantly larger problems than the current state of the art.

## 1 INTRODUCTION

Bayesian networks are a common machine learning technique used to represent relationships among variables in data sets. When these relationships are not known *a priori*, the structure of the network must be learned. A common learning approach entails searching for a structure which optimizes a particular scoring function (Cooper and Herskovits 1992; Heckerman, Geiger, and Chickering 1995). Because of the difficulty of the problem, early approaches focused on approximation techniques to learn "good" networks (Cooper and Herskovits 1992; Heckerman, Geiger, and Chickering 1995; Heckerman 1998; Friedman, Nachman, and Peer 1999; Tsamardinos, Brown, and Aliferis 2006). Unfortunately, these algorithms are unable to guarantee anything about the quality of the learned networks.

Exact dynamic programming algorithms have been developed to learn provably optimal Bayesian network structures (Ott, Imoto, and Miyano 2004; Koivisto and Sood 2004; Singh and Moore 2005; Silander and Myllymaki 2006). These algorithms identify optimal small subnetworks and add optimal leaves to find large optimal networks until finding the optimal network including all variables. Unfortunately, all of these algorithms must store an exponential number of subnetworks and associated information in memory. Parviainen and Koivisto (2009) recently proposed a divide-and-conquer algorithm in which fewer subnetworks are stored in memory at once at the expense of longer running time. Theoretical results suggest that this algorithm is slower than dynamic programming when an exponential number of processors is not available.

Yuan *et al.* (2011) developed an A* heuristic search formulation based on the dynamic programming recurrences to learn optimal network structures. The algorithm formulates the learning problem as a shortest-path finding problem in a search graph. Each path in the graph corresponds to an ordering of the variables, and each edge on the path has a cost that corresponds to the choice of an optimal parent set for one variable out of the variables that appear earlier on the path. Together, all the edges on a path encode an optimal directed acyclic graph that is consistent with the path. The solution to the shortest-path finding problem then corresponds to an optimal Bayesian network structure. The A* algorithm also uses a consistent heuristic function to prune provably suboptimal solutions during the search so as to improve its efficiency.

de Campos *et al.* (2009) proposed a systematic search algorithm to identify optimal network structures. The algorithm begins by calculating optimal parent sets for all variables. These sets are represented as a directed graph that may have cycles. Cycles are then repeatedly broken by removing one edge at a time. The algorithm terminates with an optimal Bayesian network. However, this algorithm is shown to often learn the optimal structure slower than the dynamic programming algorithm (de Campos, Zeng, and Ji 2009).

Optimal networks have also been learned using linear programming (Jaakkola et al. 2010). This technique reformu-lates the structure learning problem as a linear program. An exponential number of constraints are used to define a convex hull in which each vertex corresponds to a DAG. Coordinate descent is used to identify the vertex which corresponds to the optimal DAG structure. Furthermore, the dual of their formulation provides an upper bound which can help guide the descent algorithm. This algorithm was shown to have similar or slightly better runtime performance as dynamic programming (Jaakkola et al. 2010).

This paper describes a novel *frontier breadth-first branch and bound algorithm using delayed duplicate detection* for learning optimal Bayesian network structures. The basic idea is to formulate the learning task as a graph search problem. The search graph decomposes into natural layers and allows searching one layer at a time. This algorithm improves the scalability of learning optimal Bayesian network structures in three ways. First, the frontier search approach allows us to reduce the memory complexity by working with only a single layer of search graphs at a time during the search. In particular, we store one layer of each search graph, the scores required for that layer and information for solution reconstruction from every previous layer. Other information is deleted. In comparison, previous dynamic programming algorithms have to store an entire exponentially-sized graph in memory. Second, branch and bound techniques allow us to safely prune unpromising search nodes from the search graphs, while dynamic programming algorithms have to evaluate the whole search space. Finally, we use a delayed duplicate detection method to ensure that, given enough hard disk space, optimal network structures can be learned regardless of the amount of RAM. Previous algorithms fail if an exponential amount of RAM is not available.

The remainder of this paper is structured as follows. Section 2 provides an overview of the task of Bayesian network learning. Section 3 and 4 introduce two formulations for solving the learning task: dynamic programming and graph search. Section 5 discusses the details of the external-memory frontier breadth-first branch and bound algorithm we propose in this paper. Section 6 compares the algorithm against several existing approaches on a set of benchmark machine learning datasets. Finally, Section 7 concludes the paper.

## 2   BACKGROUND

A Bayesian network consists of a directed acyclic graph (DAG) structure and a set of parameters. The vertices of the graph each correspond to a random variable $\mathbf{V} = \{X_1, ..., X_n\}$. All parents of $X_i$ are referred to as $PA_i$. A variable is conditionally independent of its non-descendants given its parents. The parameters of the network specify a conditional probability distribution, $P(X_i|PA_i)$ for each $X_i$.

Given a dataset $\mathbf{D} = \{D_1, ..., D_N\}$, where $D_i$ is an instantiation of all the variables in $\mathbf{V}$, the optimal structure is the DAG over all of the variables which best fits $\mathbf{D}$ (Heckerman 1998). A scoring function measures the fit of a network structure to $\mathbf{D}$. For example, the minimum description length (MDL) scoring function (Rissanen 1978) uses one term to reward structures with low entropy and another to penalize complex structures. Optimal structures minimize the score. Let $r_i$ be the number of states of the variable $X_i$, let $N_{pa_i}$ be the number of data records consistent with $PA_i = pa_i$, and let $N_{x_i,pa_i}$ be the number of data records consistent with $PA_i = pa_i$ and $X_i = x_i$. The MDL score for a structure $G$ is defined as follows (Tian 2000),

$$MDL(G) = \sum_i MDL(X_i|PA_i), \quad (1)$$

where

$$\begin{aligned}
MDL(X_i|PA_i) &= H(X_i|PA_i) + \frac{\log N}{2} K(X_i|PA_i), \\
H(X_i|PA_i) &= -\sum_{x_i,pa_i} N_{x_i,pa_i} \log \frac{N_{x_i,pa_i}}{N_{pa_i}}, \quad (2) \\
K(X_i|PA_i) &= (r_i - 1) \prod_{X_l \in PA_i} r_l.
\end{aligned}$$

MDL is decomposable (Heckerman 1998), so the score for a structure is simply the sum of the score for each variable. Our algorithm can be adapted to use any decomposable function. Some sets of parents cannot form an optimal parent for any variable, as described in the following theorems from Tian (2000) and de Campos *et al.* (2009).

**Theorem 1.** *In an optimal Bayesian network based on the MDL scoring function, each variable has at most $\log(\frac{2N}{\log N})$ parents, where $N$ is the number of data points.*

**Theorem 2.** *Let $\mathbf{U} \subset \mathbf{V}$ and $X \notin \mathbf{U}$. If $BestMDL(X, \mathbf{U}) < BestMDL(X, \mathbf{V})$, $\mathbf{V}$ cannot be the optimal parent set for $X$.*

## 3   DYNAMIC PROGRAMMING

Learning an optimal Bayesian network structure is NP-Hard (Chickering 1996). Dynamic programming algorithms learn optimal network structures in $O(n2^n)$ time and memory (Ott, Imoto, and Miyano 2004; Koivisto and Sood 2004; Singh and Moore 2005; Silander and Myllymaki 2006). Because a network structure is a DAG, the optimal structure can be divided into an optimal leaf vertex and its parents as well as an optimal subnetwork for the rest of the variables. This subnetwork is also a DAG, so it can recursively be divided until the subnetwork is only a single vertex. At that point, the optimal parents have been found for all variables in the network and the optimal structure can be constructed. It has been shown (Silander and Myllymaki 2006) that a more efficient algorithm begins with a 0-variable subnetwork and exhaustively adds optimal leaves. For the MDL scoring function and variables **V**, this recurrence can be expressed as follows (Ott, Imoto, and Miyano 2004),

$$MDL(\mathbf{V}) = \min_{X \in \mathbf{V}} \{MDL(\mathbf{V} \setminus \{X\}) + BestMDL(X, \mathbf{V} \setminus \{X\})\},$$

where

$$BestMDL(X, \mathbf{V} \setminus \{X\}) = \min_{PA_X \subseteq \mathbf{V} \setminus \{X\}} MDL(X|PA_X).$$

As this recurrence suggests, all dynamic programming algorithms must perform three steps. First, they must calculate the score of each variable given all subsets of the other variables as parents. There are $n2^{n-1}$ of these scores. Then, $BestMDL$ must be calculated. For a variable $X$ and set of possible parents **V**, this function returns the subset of those parents which minimizes the score for $X$ as well as that score. There are $n2^{n-1}$ of these optimal parent sets. Finally, the optimal subnetworks must be learned. These subnetworks use $BestMDL$ to learn the optimal leaf for every possible subnetwork, including the optimal network with all of the variables. There are $2^n$ optimal subnetworks.

## 4 GRAPH SEARCH FORMULATION

We first formulate each phase of the dynamic programming algorithm as a separate search problem, including calculating parent scores, identifying the optimal parent sets, and learning the optimal subnetworks.

We use an AD-tree-like search to calculate all of the parent scores. An AD-tree (Moore and Lee 1998) is an unbalanced tree which contains AD-nodes and varying nodes. The tree is used to collect count statistics from a dataset. An AD-node stores the number of records consistent with the variable instantiation of the node, while a varying node assigns a value to a variable. As shown in Equation 2, the entropy component of a score can be calculated based on variable instantiation counts. Each AD-node has an instantiation of a set of variables **U** and the count of records consistent with that instantiation. That count is a value of $pa_i$ for all $X \in \mathbf{V} \setminus \mathbf{U}$. Furthermore, it is a value of $x_i, pa_i$ for all $X \in \mathbf{U}$ with parents $\mathbf{U} \setminus \{X\}$. We can use a depth-first traversal of the AD-tree to compute the scores. Theorem 1 states that only small parent sets can possibly be optimal parents when using the MDL score. All nodes below the depth specified in the theorem are pruned. The scores which are not pruned are written to disk for retrieval when identifying optimal parent sets. We call this data structure a *score cache*. Each entry in the score cache contains one value of $MDL(X|PA)$.

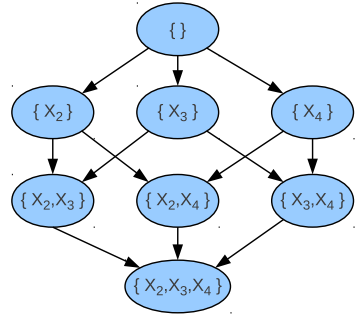

Figure 1: Parent graph for variable $X_1$

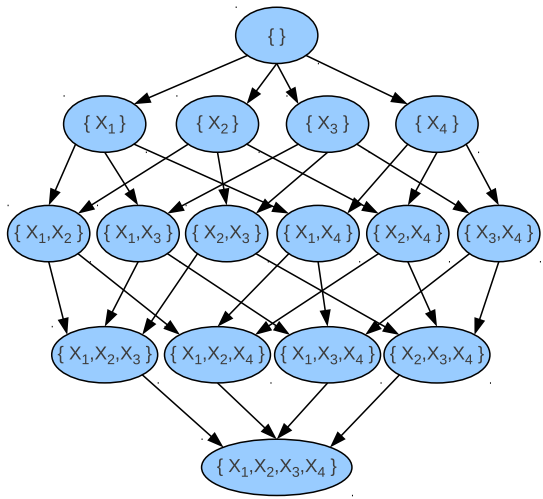

Figure 2: An order graph of four variables

A *parent graph* is a lattice in which each node stores one value of $BestMDL$ for different candidate sets of variables. The score cache is used to quickly look up the scores for candidate parent sets. Figure 1 shows the construction of the parent graph for variable $X_1$ as a lattice. All $2^{n-1}$ subsets of all other variables are present in the graph. Each node contains one value for $BestMDL$ of $X_1$ and the set of candidate parents shown. That is, each node stores the subset of parents from the given candidate set which minimizes the score of $X_1$, as well as that score. The lattice divides the nodes into layers. We call the first layer of the graph, the layer with the single node for {} in Figure 1, layer 0. A node in layer $l$ has $l$ predecessors, all in layer $l-1$, and considers candidate parent sets of size $l$. Layer $l$ has $C(n-1, l)$ nodes, where $C(n, k)$ is the binomial coefficient. Each variable has a separate parent graph. The complete set of parent graphs stores $n2^{n-1}$ parent sets.

An *order graph* is also a lattice. Each node contains $MDL(\mathbf{V})$ and the associated optimal subnetwor for one subset of variables. Figure 2 displays an order graph for four variables. Its lattice structure is similar to that of the parent graphs; because it contains subsets of all variables, though, the order graph has $2^n$ nodes. The topmost node in layer 0 containing no variables is the *start node*. The bottom-most node containing all variables is the *goal node*. A directed path in the order graph from the start node to any other node induces an ordering on the variables in the path with new variables appearing later in the ordering. For example, the path traversing nodes $\{\}, \{X_1\}, \{X_1, X_2\}, \{X_1, X_2, X_3\}$ stands for the variable ordering $X_1, X_2, X_3$. All variables which precede a variable in the ordering are candidate parents of that variable. Each edge on the path has a cost equal to $BestMDL$ for the new variable in the child node given the variables in the parent node as candidate parents. The parent graphs are used to quickly retrieve these costs. For example, the edge between $\{X_1, X_2\}$ and $\{X_1, X_2, X_3\}$ has a cost equal to $BestMDL(X_3, \{X_1, X_2\})$. Each node contains a subset of variables, the cost of the best path from the start node to that node, a leaf variable and its optimal parent set. The shortest paths from the start node to all the other nodes correspond to the optimal subnetworks, so the shortest path to the goal node corresponds to the optimal Bayesian network. The lattice again divides the nodes into layers. Nodes in layer $l$ contain optimal subnetworks of $l$ variables. Layer $l$ has $C(n,l)$ nodes.

## 5 AN EXTERNAL-MEMORY FRONTIER BREADTH-FIRST BRANCH AND BOUND ALGORITHM

Finding an optimal Bayesian network structure can be considered a search through the order graph. This formulation allows the application of any graph search algorithm, such as A* (Yuan, Malone, and Wu 2011), to find the best path from the start node to the goal node. In particular, such a formulation allows us to treat the order and parent graphs as implicit search graphs. That is, we do not have to keep the entire graphs in memory at once. Dynamic programming can be considered as a breadth-first search through this graph (Malone, Yuan, and Hansen 2011). Previous results show that the scalability of existing algorithms for learning optimal Bayesian networks is typically limited by the amount of RAM available. To eliminate the constraint of limited RAM, we introduce a *frontier breadth-first branch and bound algorithm with delayed duplicate detection* to do the search by adapting the breadth-first heuristic search algorithm proposed by Zhou and Hansen (2003; 2006). It is also similar to the frontier search described by Korf (2005).

Breadth-first heuristic search expands a search space in order of layers of increasing $g$-cost with each layer comprising all nodes with a same $g$-cost. As each node is generated, a heuristic function is used to calculate a lower bound for

**Algorithm 1** A Frontier BFBnB Search Algorithm

**procedure** EXPANDORDERGRAPH($l, isPresent, upper, lb, maxSize$)
    **for** each $MDL_l(\mathbf{U}) \in MDL_l$ **do**
        **for** each $X \in \mathbf{V} \setminus \mathbf{U}$ **do**
            $s \leftarrow MDL_l(\mathbf{U}) + BMDL_l(X|\mathbf{U}) - lb(X)$
            **if** $s > upper$ **then** continue
            $isPresent(\mathbf{U} \cup \{X\}) \leftarrow$ true
            **if** $s < MDL_{l+1}(\mathbf{U} \cup \{X\})$ **then**
                $MDL_{l+1}(\mathbf{U} \cup \{X\}) \leftarrow s$
                $MDL_{Pl+1}(\mathbf{U} \cup \{X\}) \leftarrow BMDL_{Pl}(X|\mathbf{U})$
            **end if**
            **if** $|MDL_{l+1}| > maxSize$ **then**
                writeTempFile($MDL_{l+1}, MDL_{Pl+1}$)
            **end if**
        **end for**
    **end for**
    writeTempFile($MDL_{l+1}, MDL_{Pl+1}$)
    $MDL_{l+1}, MDL_{Pl+1} \leftarrow$ mergeTempFiles
    delete $MDL_l$
**end procedure**

**procedure** EXPANDPARENTGRAPH($l, p, isPresent, maxSize$)
    **for** each $BestMDL_l(p|\mathbf{U}) \in BestMDL_l(p$ **do**
        **for** each $X \in \mathbf{V} \setminus \mathbf{U}$ and $X \neq p$ **do**
            $\mathbf{S} \leftarrow \mathbf{U} \cup \{X\}$
            **if** $!isPresent(\mathbf{S})$ **then** continue
            **if** $MDL(p|\mathbf{S}) < BMDL_{l+1}(p|\mathbf{S})$ **then**
                $BMDL_{l+1}(p|\mathbf{S}) \leftarrow MDL(p|\mathbf{S})$
                $BMDL_{Pl+1}(p|\mathbf{S}) \leftarrow \mathbf{S}$
            **end if**
            **if** $BMDL_l(p|\mathbf{U}) < BMDL_{l+1}(p|\mathbf{S})$ **then**
                $BMDL_{l+1}(p|\mathbf{S}) \leftarrow BMDL_l(p|\mathbf{U})$
                $BMDL_{Pl+1}(p|\mathbf{S}) \leftarrow BMDL_{Pl}(p|\mathbf{U})$
            **end if**
            **if** $|BMDL_{l+1}(p)| > maxSize$ **then**
                writeTempFile($BMDL_{l+1}(p), BMDL_{Sl+1}(p)$)
            **end if**
        **end for**
    **end for**
    writeTempFile($BMDL_{l+1}(p), BMDL_{Pl+1}(p)$)
    $BMDL_{l+1}, BMDL_{Pl+1}(p) \leftarrow$ mergeTempFiles
    delete $BMDL_l, BMDL_{Pl}(p)$
**end procedure**

**procedure** EXPANDADNODE($i, \mathbf{U}, \mathbf{D}_u, d$)
    **For** $j = i + 1 \rightarrow n$ **do** expandVaryNode($j, \mathbf{U}, \mathbf{D}_u, d$)
**end procedure**

**procedure** EXPANDVARYNODE($i, \mathbf{U}, \mathbf{D}_u, d$)
    **for** $j = 0 \rightarrow r_i$ **do**
        updateScores($\mathbf{U} \cup \{X_i\}, \mathbf{D}_{X_i=j,u}$)
        **if** $d > 0$ **then** expandADNode($i, \mathbf{U} \cup \{X_i\}, \mathbf{D}_{X_i=j,u}, d-1$)
    **end for**
**end procedure**

**procedure** UPDATESCORES($\mathbf{U}, \mathbf{D}_u$)
    **for** $X \in \mathbf{V} \setminus \mathbf{U}$ **do**
        **if** $MDL(X|\mathbf{U})$ is null **then** $MDL(X|\mathbf{U}) \leftarrow K(X|\mathbf{U})$
        $MDL(X|\mathbf{U}) \leftarrow MDL(X|\mathbf{U}) + N_u * \log N_u$
    **end for**
    **for** $X \in \mathbf{U}$ **do**
        **if** $MDL(X|\mathbf{U} \setminus \{X\})$ is null $MDL(X|\mathbf{U} \setminus \{X\}) \leftarrow K(X|\mathbf{U} \setminus \{X\})$
        $MDL(X|\mathbf{U} \setminus \{X\}) \leftarrow MDL(X|\mathbf{U} \setminus \{X\}) - N_u * \log N_u$
    **end for**
**end procedure**

**procedure** MAIN($\mathbf{D}, upper, maxSize$)
    $maxParents \leftarrow \log \frac{2N}{\log N}$
    expandADNode($-1, \{\}, \mathbf{D}, maxParents$)
    $lb \leftarrow$ getBestScores
    writeScoresToDisk
    $isPresent \leftarrow \{\}$
    **for** $l = 1 \rightarrow n$ **do**
        **for** $p = 1 \rightarrow n$ **do**
            expandParentGraph($l, p, isPresent, maxSize$)
        **end for**
        expandOrderGraph($l, isPresent, upper, lb, maxSize$)
    **end for**
    $optimalStructure \leftarrow$ reconstructSolution
**end procedure**

that node. If the lower bound is worse than a given upper bound on the optimal solution, the node is pruned; otherwise, the node is added to the open list for further search. After the search, a divide-and-conquer method is used to reconstruct the optimal solution.

Algorithm 1 gives the pseudocode for our BFBnB search algorithm for learning optimal Bayesian networks. The algorithm is very similar to the breadth-first heuristic search algorithm but has several subtle and important differences. First, the layers in our search graphs (the parent and order graphs) do not correspond to the $g$-costs of nodes; rather, layer $l$ corresponds to variable sets (candidate parent sets or optimal subnetworks) of size $l$. For the order graph, though, we can calculate both a $g$- and $h$-cost for pruning, as described in Section 5.1. We also describe how to propagate this pruning from the order graph to the parent graphs. Another difference is that our search problem is a nested search of order and parent graphs. The layered parent and order graph searches have to be carefully orchestrated to ensure the correct nodes can be accessed easily at the correct time, as described in Section 5.2. This further requires the parent scores are stored in particular order, as described in Section 5.3. Yet another difference is that we use a variant of delayed duplicate detection (Korf 2008) in which a hash table is used to detect as many duplicates in RAM as possible before resorting to external memory, as described in Section 5.4. Finally, we store a portion of each order graph node to reconstruct the optimal network structure after the search, as described in Section 5.5.

## 5.1 BRANCH AND BOUND

We need a heuristic function $f(\mathbf{U}) = g(\mathbf{U}) + h(\mathbf{U})$ that estimates the cost of the best path from the start node to a goal node using order node $\mathbf{U}$. The $g$ cost is simply the sum of the edge costs of the best path from the start node to $\mathbf{U}$. The $h$ cost provides a lower bound on the cost from $\mathbf{U}$ to the goal node. We use the following heuristic function $h$ from Yuan *et al.* (2011).

**Definition 1.**
$$h(\mathbf{U}) = \sum_{X \in \mathbf{V} \setminus \mathbf{U}} BestMDL(X, \mathbf{V} \setminus \{X\}). \quad (3)$$

This heuristic function relaxes the acyclic constraint on the remaining variables in $\mathbf{V} \setminus \mathbf{U}$ and allows them to choose parents from all of the variables in $\mathbf{V}$. The following theorem from Yuan *et al.* (2011) proves that the function is consistent. Consistent heuristics are guaranteed to be admissible.

**Theorem 3.** *h is consistent.*

In order to calculate this bound, we must know $BestMDL(X, \mathbf{V} \setminus \{X\})$. Fortunately, these scores are calculated during the first phase of the algorithm. Because the score cache contains every score which could possibly be optimal for all variables, it is guaranteed to have the optimal score for all variables given any set of parents, which is $BestMDL(X, \mathbf{V} \setminus \{X\})$. Thus, we can identify these scores while calculating the scores when expanding the AD-tree and store them in an array for reuse. The pseudocode uses the function $getBestScores$ to find these scores and the array $lb$ to store them.

We can apply BFBnB to prune nodes in the order graph using the lower bound function in Equation 3; however, pruning is not directly applicable to the parent graphs. An optimal parent score $BestMDL(X, \mathbf{U})$ is only necessary if a node for $\mathbf{U}$ is in the order graph. Consequently, if $\mathbf{U}$ is pruned from the order graph, then the nodes for $\mathbf{U}$ are also pruned from the parent graphs. The pseudocode uses $isPresent$ to track which nodes were not pruned.

We also need an upper bound score on the optimal Bayesian network for pruning. A search node $\mathbf{U}$ whose heuristic value $f(\mathbf{U})$ is higher than the upper bound is immediately pruned. Numerous fast, approximate methods exist for learning a locally optimal Bayesian network. We use a greedy beam search algorithm based on a local search algorithm described by Heckerman (1998) to quickly find the upper bound. A more sophisticated algorithm could be used to find a better bound and improve pruning. The input argument $upper$ is this bound in the pseudocode.

## 5.2 COORDINATING THE GRAPH SEARCHES

The parent and order graph searches must be carefully coordinated to ensure that the parent graphs contain the necessary nodes to expand nodes in the order graph. In particular, expanding a node $\mathbf{U}$ in layer $l$ in the order graph requires $BestMDL(X, \mathbf{U})$, which is stored in the node $\mathbf{U}$ of the parent graph for $X$. Hence, before expanding layer $|\mathbf{U}|$ in the order graph, that layer of the parent graphs must already exist. Therefore, the algorithm alternates between expanding layers of the parent graphs and order graph.

Expanding a node $\mathbf{U}$ in the parent graph amounts to generating successor nodes with candidate parents $\mathbf{U} \cup \{X\}$ for all $X$ in $\mathbf{V} \setminus \mathbf{U}$. For each successor $\mathbf{S} = \mathbf{U} \cup \{X\}$, the hash table for the next layer is first checked to see if $\mathbf{S}$ has already been generated. If not, the score of using all of $\mathbf{S}$ as parents of $X$ is retrieved from the score cache and compared to the score of using the parents specified in $\mathbf{U}$. If using all of the variables has a better score, then an entry is added to the hash table indicating that, for possible parents $\mathbf{S}$, using all of them is best. Otherwise, according to Theorem 2, the hash table stores a mapping from $\mathbf{S}$ to the parents in $\mathbf{U}$. Similarly, if $\mathbf{S}$ has already been generated, the score of the existing best parent set for $\mathbf{S}$ is compared to the score using the parents in $\mathbf{U}$. If the score of the parents in $\mathbf{U}$ is better, then the hash table mapping is updated accordingly. Once a layer of the parent graph is expanded, the whole layer can be discarded as it is no longer needed. The pseudocode uses $BMDL_l$ to store the optimal scores and $BMDL_{Pl}$ to store the optimal parents.

Expanding a node $\mathbf{U}$ in the order graph amounts to generating successor nodes $\mathbf{U} \cup \{X\}$ for all $X$ in $\mathbf{V} \setminus \mathbf{U}$. To calculate the score of successor $\mathbf{S} = \mathbf{U} \cup \{X\}$, the score of the existing node $\mathbf{U}$ is added to $BestMDL(X, \mathbf{U})$, which is retrieved from parent graph node $\mathbf{U}$ for variable $X$. The optimal parent set out of $\mathbf{U}$ is also recorded. This is equivalent to trying $X$ as the leaf and $\mathbf{U}$ as the subnetwork. Next, the hash table for the next layer is consulted. If it contains an entry for $\mathbf{S}$, then a node for this set of variables has already been generated using another variable as the leaf. The score of that node is compared to the score for $\mathbf{S}$. If the score for $\mathbf{S}$ is better, or the hash table did not contain an entry for $\mathbf{S}$, then the mapping in the hash table is updated. Unlike the parent graph, however, a portion of each order graph node is used to reconstruct the optimal network at the end of the search, as described in Section 5.5. This information is written to disk, while the other information is deleted. The pseudocode uses $MDL_l$ to store the score for each subnetwork and $MDL_{Pl}$ to store the associated parent information.

Additional care is needed to ensure that parent and order graph nodes for a particular layer are accessed in a regular, structured pattern. We arrange the nodes in the parent and order graphs in queues such that when node $\mathbf{U}$ is removed from the order graph queue, the head of each parent graph queue for all $X$ in $\mathbf{V} \setminus \mathbf{U}$ is $\mathbf{U}$. So all of the successors of $\mathbf{U}$ can be generated by combining it with the head of each of those parent graph queues. Once the parent graph nodes are used, they can be removed, and the queues will be ready to expand the next node in the order graph queue. Because the nodes are removed from the heads of the queues, these invariants hold throughout the expansion of the layer. Regulating such access patterns improves the scalability of the algorithm because these queues can be stored on disk and accessed sequentially to reduce the requirement of RAM. The regular accesses also reduce disk seek time. The pseudocode assumes the nodes are written to disk in this order to easily retrieve the next necessary node.

The lexicographic ordering (Knuth 2009) of nodes within each layer is one possible ordering that ensures the queues remain synchronized. For example, the lexicographic ordering of 4 variables of size 2 is $\{\{X_1, X_2\}, \{X_1, X_3\}, \{X_2, X_3\}, \{X_1, X_4\}, \{X_2, X_4\}, \{X_3, X_4\}\}$. The order graph queue for layer 2 of a dataset with 4 variables should be arranged in that order. The parent graph queue for variable $X$ should have the same sequence, but without subsets containing $X$. In the example, the parent graph queue for variable $X_1$ should be $\{\{X_2, X_3\}, \{X_2, X_4\}, \{X_3, X_4\}\}$. As described in more detail in Section 5.4, the nodes of the graphs must be sorted to detect duplicates; the lexicographic order ensures that there is no additional work required to arrange the nodes when writing them to disk.

### 5.3 ORDERING THE SCORES ON DISK

For large datasets, the score cache can grow quite large. We write it to disk to reduce RAM usage. Each score $MDL(X, \mathbf{U})$ is used once, when node $\mathbf{U}$ is first generated in the parent graph for $X$. As described in Section 5.2, the parent graph nodes are expanded in lexicographic order; however, they are not generated in that order. The score $MDL(X, \mathbf{U}), \mathbf{U} = \{Y_1 \ldots Y_l\}$ is needed when expanding node $\mathbf{U} \setminus \{Y_l\}$ in the parent graph for $X$. Therefore, the scores must be written to disk in that order. The pseudocode uses the $writeScoresToDisk$ function to sort and write the scores to disk in this order.

A file is created for each variable for each layer to store these sorted scores. The file for a particular layer can be deleted after expanding that layer in the appropriate parent graph.

### 5.4 DUPLICATE DETECTION

Duplicate nodes are generated during the graph searches. Duplicates in the parent and order graphs correspond to nodes which consider the same sets of variables (candidate parent sets and optimal subnetworks, respectively). Because the successors of a node always consider exactly one more variable in both the parent and order graphs, the successors of a node in layer $l$ are always in layer $l + 1$. Therefore, when a node is generated, it could only be a duplicate of a node in the open list for layer $l + 1$. In both the parent and order graphs, the duplicate with the best score should be kept.

For large datasets, it is possible that even one layer of the parent or order graph is too large to fit in RAM. We use a variant of the *delayed duplicate detection* (DDD) (Korf 2008) in our algorithm to utilize external memory to solve such large learning problems. In DDD, search nodes are written to a file on disk as they are generated. After expanding a layer, an external-memory sorting algorithm is used to detect and remove duplicate nodes in the file. The nodes in the file are then expanded to generate the next layer of the search. Consequently, the search uses a minimal amount of RAM; however, all generated nodes are written to disk, so much work is done reading and writing duplicates.

Rather than immediately writing all generated nodes to disk, we instead detect duplicates in RAM as usual with a hash table. Once the open list reaches a user-defined maximum size, its contents are sorted and written to a temporary file on disk. The open list is then cleared. At the end of each layer, the remaining contents of the open list and the temporary files are sorted and merged into a single file which contains the sorted list of nodes from that layer. For rea-sons described in Section 5.2, the lexicographic ordering of nodes within a layer is used when sorting. The hash table reduces the number of nodes written to and read from disk by detecting as many duplicates as possible in RAM.

The pseudocode uses $maxSize$ as the user-defined maximum size. The function $writeTempFile$ sorts, writes to disk and clears the open list provided as its argument. The scores and optimal parent sets are written together on disk. The function $mergeTempFiles$ performs an external memory merge to detect duplicates in the temp files. For the parent graphs, both the scores and optimal parent sets are kept in a single file; however, as described in Section 5.5, the parent information of the order graph must be stored for the entire search, while the score information can be deleted after use. Therefore, two separate files are used to allow the information to easily be deleted.

## 5.5 RECONSTRUCTING THE OPTIMAL NETWORK STRUCTURE

In order to trace back the optimal path and reconstruct the optimal network structure, we write a portion of each node of the order graph to a disk file once it is expanded during the order graph search. For each order graph node we write the subset of variables, the leaf variable and its optimal parents. Solution reconstruction works as follows. The final leaf variable $X$ and its optimal parent set are retrieved from the goal node. Because the goal node considers all variables, its predecessor in the optimal path is $\mathbf{U} = \mathbf{V} \setminus \{X\}$. This predecessor is retrieved from the file for layer $|\mathbf{U}|$. That node has the optimal leaf and parent set for that subnetwork. Recursively, the optimal leaves and parent sets are retrieved until reconstructing the entire network structure. We use this approach instead of the standard divide-and-conquer solution reconstruction because, as shown in Section 6, it requires relatively little memory. Furthermore, divide-and-conquer would require regeneration of the parent graphs, which is quite expensive in terms of time and memory. The pseudocode uses the function $reconstructSolution$ to extract this information from the $MDL_{Pl}$ files.

## 5.6 ADVANTAGES OF OUR ALGORITHM

Our frontier breadth-first branch and bound algorithm has several advantages over previous algorithms for learning optimal Bayesian networks.

First, our top-down search of the AD-tree for calculating scores ensures we never need to calculate scores or counts of large variable sets. The AD-tree method is in contrast to the bottom-up method used by other algorithms (Silander and Myllymaki 2006). Bottom-up methods must always compute the scores, or at least the counts, of large parent sets in order to correctly calculate the counts required for the smaller ones. Since our algorithm neither calculates nor stores these counts and scores, it both runs more quickly and uses less memory.

Second, the layered search strategy reduces the memory requirements by working with one layer of the parent and order graphs at a time. Other information can be either discarded immediately or stored in hard disk files for later use, e.g., the information needed to reconstruct the optimal network structure. Previous formulations, such as P-Caches (Singh and Moore 2005) and arrays (Silander and Myllymaki 2006), could not take advantage of this structure. Singh and Moore propose a depth-first search through the P-Caches, while Silander and Myllymaki's approach identifies the sets according to their lexicographic ordering. (We use the lexicographic order within each layer, not over all of the variables.) These approaches can identify neither optimal parent sets nor optimal subnetworks one layer at a time. Thus, they must both keep all of the optimal parent sets and subnetworks in memory.

Third, we prune the order graph using an admissible heuristic function; this further reduces the memory complexity of the algorithm. Pruning unpromising nodes from the order graph not only reduces the amount of computation but also reduces the memory requirement. Furthermore, the savings in running time and memory also propagate to parent graphs. Dynamic programming algorithms always evaluate the full order graph.

The duplicate detection method we use lifts the requirement that open lists fit in RAM to detect duplicates. Because our algorithm does not resort to delayed duplicate detection until RAM is full, our algorithm can still take advantage of large amounts of RAM. By writing nodes to disk, we can learn optimal Bayesian networks even when single layers of the search graphs do not fit in RAM.

Our algorithm also has advantages over other learning formulations. In contrast to the A* algorithm of Yuan *et al.*( 2011), we only keep one layer of the order graph in memory at a time. The open and closed lists of A* keep all generated nodes in memory to perform duplicate detection. Unlike the systematic search algorithm of de Campos *et al.* (de Campos, Zeng, and Ji 2009), we always search in the space of DAGs, which is smaller than the space of directed graphs in which that algorithm searches. The LP algorithm (Jaakkola et al. 2010) uses the same mechanism to identify optimal parent sets as DP; therefore, it cannot complete when all optimal parent sets do not fit in memory.

## 6 EXPERIMENTS

We compared a Java implementation of the external-memory frontier BFBnB search with DDD (BFBnB) to an efficient version (Silander and Myllymaki 2006) of dynamic programming which uses external memory written in C downloaded from http:/b-course.hiit.fi/bene. We refer to it as DP. Previous results (Silander and Myllymaki 2006) have shown DP is more efficient than other dynamic programming implementations. We also compared to Yuan et al.'s A* implementation (2011) (A*) and de Campos et al.'s branch and bound systematic search algorithm (de Campos, Zeng, and Ji 2009) (SS) downloaded from http://www.ecse.rpi.edu/ cvrl/structlearning.html. We did not include comparison to the DP implementation of Malone et al. (2011) (MDP) because the codebase is similar; however, MDP does not incorporate pruning or delayed duplicate detection. The running times of BFBnB and MDP are similar on datasets which both complete, but, due to duplicate detection, MDP fails when an entire layer of the order graph does not fit in RAM.

| Dataset | | | Timing Results (s) | | | | Space Results (bytes) | |
|---|---|---|---|---|---|---|---|---|
| dataset | n | N | DP | BFBnB | A* | SS | DP | BFBnB |
| wine | 14 | 178 | 1 | 0 | 0 | 171 | 1.16E+07 | 2.72E+05 |
| adult | 14 | 30,162 | 1 | 18 | 11 | OT | 1.16E+07 | 1.36E+06 |
| zoo | 17 | 101 | 1 | 1 | 0 | OT | 4.81E+07 | 2.30E+06 |
| houseVotes | 17 | 435 | 7 | 5 | 3 | 5,824 | 4.81E+07 | 4.39E+06 |
| letter | 17 | 20,000 | 29 | 87 | 116 | OT | 4.81E+07 | 9.12E+06 |
| statlog | 19 | 752 | 23 | 9 | 12 | OT | 1.82E+08 | 1.82E+07 |
| hepatitis | 20 | 126 | 27 | 9 | 6 | 202 | 3.79E+08 | 2.73E+07 |
| segment | 20 | 2,310 | 44 | 28 | 42 | 2,482 | 3.79E+08 | 3.67E+07 |
| meta | 22 | 528 | 52 | 57 | 41 | OT | 1.67E+09 | 1.55E+08 |
| imports | 22 | 205 | 123 | 54 | 55 | 3,723 | 1.67E+09 | 1.52E+08 |
| horseColic | 23 | 300 | 468 | 93 | 117 | 1,410 | 3.48E+09 | 2.41E+08 |
| spect (heart) | 23 | 267 | 413 | 131 | 139 | OT | 3.48E+09 | 3.06E+08 |
| mushroom | 23 | 8,124 | 438 | 372 | 508 | OT | 3.48E+09 | 3.14E+08 |
| parkinsons | 23 | 195 | 297 | 103 | 130 | OT | 3.48E+09 | 2.63E+08 |
| sensorReadings | 25 | 5,456 | 12,747 | 3,061 | OM | OT | 1.51E+10 | 1.30E+09 |
| autos | 26 | 159 | 2,737 | 1,184 | OM | OT | 3.15E+10 | 2.19E+09 |
| horseColic (full) | 28 | 300 | 30,064 | 4,251 | OM | OT | 1.36E+11 | 1.09E+10 |
| steelPlatesFaults | 28 | 1,941 | 78,487 | 9,252 | OM | OT | 1.36E+11 | 1.09E+10 |
| flag | 29 | 194 | 41,733 | 12,935 | OM | OT | 2.81E+11 | 1.55E+10 |
| wdbc | 31 | 569 | OD | 93,682 | OM | OT | OD | 6.86E+10 |
| epigenetic | 33 | 72,228 | OD | 570,760 | OM | OT | OD | 2.74E+11 |

Table 1: A comparison of the running time (in seconds) for Silander and Myllymaki's dynamic programming implementation (DP), Yuan et al.'s A* algorithm (A*), de Campos et al.'s systematic search algorithm (SS) and our external-memory frontier breadth-first branch and bound algorithm (BFBnB). The run times are given for all algorithms. Maximum external memory usage is given for DP and BFBnB. For reference, 1E+09 is 1 gigabyte. 'n' is the number of variables. 'N' is the number of records. 'OT' means failure to find optimal solutions due to running for more than 2 hours (7,200 seconds, less than 25 variables) or 24 hours (86,400 seconds, 25 - 29 variables) and not producing a provably optimal solution. 'OM' means failure to find optimal solutions due to running out of RAM (16GB). 'OD' means failure to find optimal solutions due to running out of hard disk space (500GB).

Benchmark datasets from the UCI repository (Frank and Asuncion 2010) were used to test the algorithms. We also constructed a biological dataset consisting of ChIP-Seq data for epigenetic features downloaded from http://dir.nhlbi.nih.gov/papers/lmi/epigenomes/hgtcell.html and http://dir.nhlbi.nih.gov/papers/lmi/epigenomes/hgtcellacetylation.aspx. The experimental datasets were normalized using linear regression using the IgG control dataset downloaded from http://home.gwu.edu/~wpeng/Software.htm. The largest datasets in the comparison have up to 33 variables and over 70,000 records. Continuous and discrete variables with more than four states were discretized into two states around the mean. Records with missing values were removed.

DP and SS do not calculate the MDL score for a network; however, they can calculate BIC. The score uses an equivalent calculation as MDL, so the algorithms always learned equivalent networks. The experiments were performed on a 3.07 GHz Intel i7 with 16GB of RAM, 500GB of hard disk space and running Ubuntu version 10.10. On datasets with less than 25 variables, all algorithms were given a maximum runtime of 2 hours (7,200 seconds). On datasets with 25 to 29 variables, all algorithms were given a maximum runtime of 24 hours (86,400 seconds).

We empirically evaluated the algorithms for both space and time requirements. For the algorithms which used external memory (BFBnB and DP), we compared the maximum hard disk usage. We also compared the running times of the algorithms. The results are given in Table 1.

Previous results found that memory is the main bottleneck restricting the size of learnable networks (Parviainen and Koivisto 2009). As the results show, algorithms which attempt to store entire parent or order graphs in RAM, such as A* and SS, are limited to smaller sets of variables. BF-BnB's duplicate detection strategy allows it to write parital search layers to hard disk when the layers are too large to fit in RAM, so it can learn optimal Bayesian network structures regardless of the amount of RAM. Consequently, hard disk space is its only memory limitation. The inexpensive cost of hard disks coupled with distributed file systems can potentially erase the effect of memory on the scalability of the algorithm.

For the datasets which it could solve, A* was sometimes faster than the other algorithms. This is unsurprising since it uses only RAM; however, it is unable to solve the larger datasets that cannot fit entirely in RAM. Even on many of the smaller datasets, though, A* runs more slowly than BF-BnB because it has the overhead cost to keep its open list in sorted order.

BFBnB not only takes an order of magnitude less external memory, but runs several times faster than the DP algorithm on most of the datasets. DP is faster on the *adult*, *letter* and *meta* datasets. These datasets have a small number of variables and a large number of records. The large number of records limits the pruning of the AD-tree from Theorem 1 and increases the runtime of BFBnB. However, BFBnB runs faster on both *mushroom* (8,000 records) and *sensorReadings* (5,000 records). Therefore, as the number of variables increases, the number of records impacts the runtime less.

The SS algorithm ran much more slowly than the other algorithms. It searches in the space of directed graphs rather than DAGs. These results suggest that search in the space of DAGs is more efficient than the space of directed graphs.

To demonstrate that our algorithm is applicable to larger datasets, we also tested it using the *wdbc* dataset (31 variables, 569 records) and a biological dataset (33 variables, 72,228 records), *epigenetic*. We learned the optimal network for *wdbc* in 93,682 seconds (about 26 hours) and the optimal network for *epigenetic* in 570,760 seconds (about 6 days). We also attempted to use DP, but its hard disk usage exceeded the 500GB of free hard disk space on the server. Figure 3 shows the total memory consumption of our algorithm for *wdbc*. Very little memory is used before layer 9, and after layer 22, the memory consumption does not change much because the layer sizes decrease. As the figure shows, both of the middle layers use nearly 70 gigabytes of disk space. Most of this space is consumed by the parent graphs, so it is is freed after each layer. Assuming that the running time and size of the middle layers double for each additional variable, which is a rough pattern from Table 1, our algorithm could learn a 36-variable network in about 50 days using approximately 2 terabytes of hard disk space and a single processor. This suggests that our method should scale to larger networks better than the method of Parviainen and Koivisto (2009). They observe that their implementation would take 4 weeks on 100 processors to learn a 31-variable network, and, even with coding improvements and massive parallelization, only networks up to 34 variables would be possible.

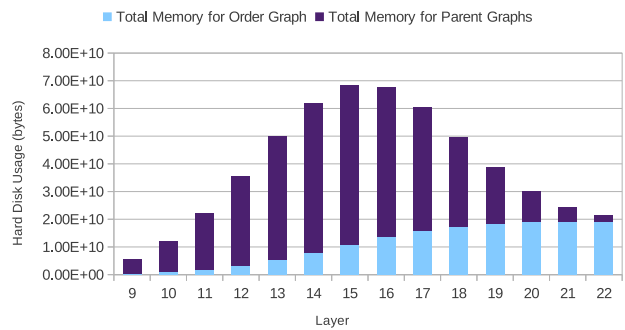

Figure 3: Hard disk usage for the wdbc dataset

## 7   CONCLUSION

Learning optimal Bayesian network structures has been thought of in terms of dynamic programming; however, such a formulation naively requires $O(n2^n)$ memory. Other formulations have been shown to have similar or slower runtimes or require other exponential resources, such as processors. This paper formulates the structure learning problem as a frontier breadth-first branch and bound search. The layered search technique allows us to work with one layer of the score cache, parent and order graphs at a time. Consequently, we delete layers of the parent graphs after expanding them and store only a portion of each order graph node to hard disk files to reduce the memory complexity. The delayed duplicate detection strategy further improves the scalability of the algorithm by writing partial layers to disk rather than requiring an entire layer fit in RAM at once. Additionally, a heuristic function allows parts of the order graph to be ignored entirely; this also reduces memory complexity and improves scalability.

Experimental results demonstrate that this algorithm outperforms the previous best implementation of dynamic programming for learning optimal Bayesian networks. Our algorithm not only runs faster than the existing approach, but also takes much less space. The LP formulation exhibits similar runtime behavior as DP, so our algorithm should similarly outperform it. It also scales to more variables than A*. Additionally, by searching in the space of DAGs instead of the space of directed graphs with cycles, it proves the optimality of the learned network more quickly than SS.

Future work will investigate better upper bounds and heuristic functions to further increase the size of learnable optimal networks. Also, like existing methods (Parviainen and Koivisto 2009; Silander and Myllymaki 2006), our algorithm can benefit from parallel computing. In addition, distributed computing can scale up our algorithm to even larger learning problems. Networks learned from our algorithm could also be used as a "gold standard" in studying the assumptions of approximate structure learning algorithms.

**Acknowledgements** This work was supported by NSF CAREER grant IIS-0953723 and EPSCoR grant EPS-0903787.